# Hierarchical Large Language Models in Cloud-Edge-End Architecture for Heterogeneous Robot Cluster Control

Zhirong Luan[1*], Yujun Lai[1], Rundong Huang[1]，Yan Yan[2]，Jingwei Wang[2]，Jizhou Lu[2]，Badong Chen[3]

[1]School of Electrical Engineering, Xi'an University of Technology, Xi'an, China

[2]China Academy of Aerospace Science and Innovation, Bei'jing, China

[3]College of Artificial Intelligence ,Xi'an Jiaotong University,Xi'an, China

E-mail:luanzhirong@xaut.edu.cn

**Abstract**. Despite their powerful semantic understanding and code generation capabilities, Large Language Models (LLMs) still face challenges when dealing with complex tasks. Multi-agent strategy generation and motion control are highly complex domains that inherently require experts from multiple fields to collaborate. To enhance multi-agent strategy generation and motion control, we propose an innovative architecture that employs the concept of a cloud-edge-end hierarchical structure. By leveraging multiple large language models with distinct areas of expertise, we can efficiently generate strategies and perform task decomposition. Introducing the cosine similarity approach,aligning task decomposition instructions with robot task sequences at the vector level, we can identify subtasks with incomplete task decomposition and iterate on them multiple times to ultimately generate executable machine task sequences. With this architecture, we successfully address the challenges of multi-agents executing open tasks in open scenarios and the task decomposition problem.

## 1 INTRODUCTION

With the rapid evolution of Artificial Intelligence (AI) technology, Large Language Models (LLMs) have made significant strides across various domains, offering new possibilities for addressing complex problems. In the field of robotics, the application of LLMs holds immense promise.

Although large language modeling in robotics has received a lot of attention, it still faces many technical challenges. Firstly, conventional approaches often employ a single LLM for robot control, leading to over-generalization and hindering the efficient execution of critical tasks such as robot strategy generation, task decomposition, and program generation [1]. Secondly, the complexity of task decomposition makes it difficult to break down complex tasks into sufficiently granular subtasks, resulting in some subtasks that robots cannot execute, necessitating further decomposition to meet execution requirements. Additionally, code generated by LLMs typically requires human intervention to transform it into executable robot instructions, introducing additional complexity and time consumption in practical applications [2].

This work introduces an innovative cloud-edge-device architecture designed to harness the capabilities of Large Language Models (LLMs) while addressing the challenge of inadequate task decomposition in the field of robotics.

The architecture comprises three essential components: the cloud, the edge, and the device. In the cloud, a powerful LLM is deployed, responsible for generating overall coordination policies for multiple robots as well as individual policies for each robot based on task requirements. At the edge, two locally deployed key models come into play: the Visual Large Model and the Linguistic Large Model. The Visual Large Model collects information about the local environment, while the Linguistic Large Model is responsible for comprehending the policies generated in the cloud and translating them into executable instruction sequences. These sequences correspond to various robot executable functions encapsulated as atomic operations, adaptable to different task decomposition subtask sequences [3].

Simultaneously, at the edge, the completeness of the task decomposition is evaluated using a cosine similarity method to ensure alignment with existing functionality. If the task decomposition results fall short of expectations, the Edge's Visual Large Model and Linguistic Large Model can be consulted to determine how to combine existing atomic tasks to achieve the desired outcome. Ultimately, the output is transmitted to the robot on the device side as a cue, ensuring comprehensive task decomposition and efficient execution of robot actions [4]. This innovative architecture is poised to excel in robot collaboration, unlocking the full potential of LLMs to address the challenge of inadequate task decomposition.

This paper is structured as follows: Section 2 discusses prior work related to LLMs inference. In Section 3, we provide a detailed description of our proposed method. Section 4 presents the experiments we conducted on multi-robot strategy generation and motion control, along with an in-depth analysis of the results. Finally, Section 5 outlines the conclusions drawn from this study.

## 2 RELATED WORK

The development and applications of Large Language Models (LLMs) have garnered significant attention across various research domains. To better understand the capabilities and applications of LLMs, we draw upon pertinent prior research contributions:

Huang et al. focused on enhancing robot comprehension of linguistic instructions by employing LLMs to generate complex trajectories for diverse manipulation tasks [5]. This approach enables efficient planning and execution of tasks in dynamic environments based on open-set instructions and objects. Wu et al. personalized household cleanup with robots, enabling them to adapt to user preferences for object placement through language-based planning and perception. Their method, facilitated by LLMs, achieved high accuracy in object placement and demonstrated real-world success with the TidyBot mobile manipulator [6]. Vemprala et al. showcased the versatility of ChatGPT in solving robotic engineering tasks via natural language commands. They also introduced a tool for collaborative cueing schemes in robotic applications, offering insights into leveraging LLMs for solving complex problems [7]. Dong et al. proposed a method for self-collaborative code generation using ChatGPT, highlighting the potential of language models in collaborative code generation efforts [8]. Shah et al. introduced LM-Nav, a novel approach to robot navigation that facilitates natural and efficient communication between robots and humans. This approach combines language models with pre-trained navigation and image-language association models, eliminating the need for costly linguistic annotations [9].These notable contributions underscore the diverse applications and potential of LLMs in the field of robotics, offering valuable insights and paving the way for our own work.

## 3 APPROACH

### 3.1 Architecture

Our proposed novel architecture aims to enable policy generation and motion control for large language models and multiple robots, addressing the issues of incomplete task decomposition in large language models and the incompatibility of generated code for direct use. The architecture is divided into three main components: cloud, edge, and device, allowing for the realization of complex multi-robot policy generation and motion control through the collaboration of multiple large language models.

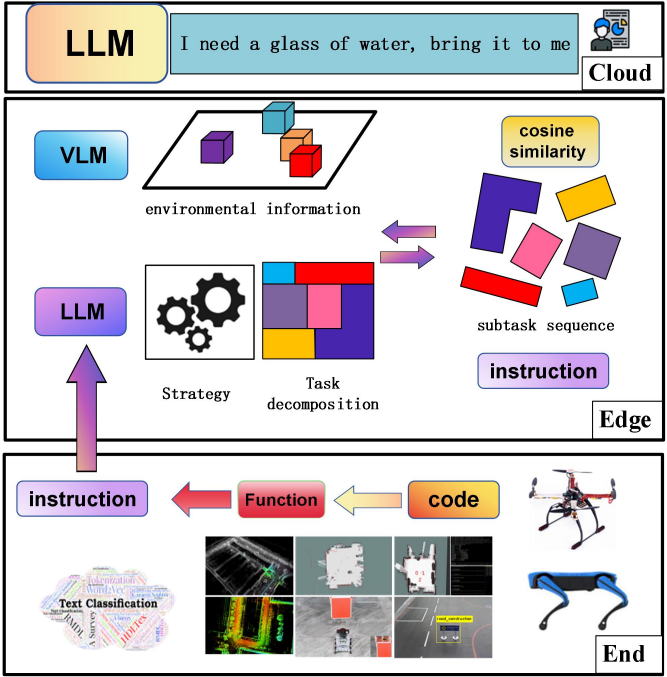

**Figure 1.** Schematic of heterogeneous unmanned cluster control based on large-scale language model in cloud-edge-end architecture.

As shown in Fig. 1 firstly in the cloud, we deployed a powerful Large Language Model (LLM), the LLMS in the cloud is mainly responsible for policy generation after receiving the task requirements, on one hand, it generates the overall cooperation policy of multiple robots to coordinate the actions of multiple robots. On the other hand, it generates strategies for individual robots to realize the cooperative work and task division of multiple robots.

At the edge, we deploy two localized key models, a visual grand model as well as a large linguistic model, where the visual grand model is responsible for acquiring local environment information, including object locations, scene information, etc. The visual grand model is used to understand the policies generated in the cloud to generate a series of sub-task sequences and transform them into executable instructions. The linguistic micromodel is used to understand the cloud-generated policies, task decompose the policies to generate a series of sub-task sequences and transform them into executable instruction sequences. These two models collaborate with each other to perform task decomposition based on the local environment information to generate instruction sequences for the robot on the device side [10].

### 3.2 Cosine similarity

Our proposed method focuses on the completeness of task decomposition by encapsulating the original functionality of the robot on the device side to ensure that the generated tasks are all pre-existing functionality, and we ensure that the tasks generated by task decomposition must be pre-existing encapsulated functionality by determining whether the task decomposition is in place using the cosine similarity method. Specifically, we compare the degree of similarity between tasks generated by task decomposition and pre-existing functions by calculating the cosine similarity. If the similarity is not high, it means that the task decomposition is not complete enough, and the undecomposed subtasks need to be re-task decomposed.

In our study, we employ a cosine similarity-based approach to determine the adequacy of task decomposition in order to optimize the depth and accuracy of task decomposition. This approach allows us to quantify the similarity between subtasks and decide whether further decomposition is required, and this decision can be derived from the following mathematical formula:

First, we represent each subtask $S_i$ as a vector containing features such as the description, keywords, contextual information of the task. Then, we calculate the cosine similarity between each pair of sub-tasks and obtain a similarity matrix. $S_{similarity}$ This matrix is represented by the following equation:

$$S_{\text{similarity}}[i][j] = \frac{S_i \cdot S_j}{\|S_i\| \cdot \|S_j\|}, i,j = 1,2,...,n \quad (1)$$

where $S_i \cdot S_j$ denotes the dot product of the vectors of subtask i and subtask j, and $\|S_i\|$ and $\|S_j\|$ denote their vector paradigms (lengths), respectively. This similarity matrix reflects the degree of similarity between subtasks.

Next, we introduced a similarity threshold of $T_{similarity}$ as a key criterion for decision making. If $S_{similarity}[i][j]$ is less than $T_{similarity}$, it indicates that the similarity between subtask i and subtask j is low and further decomposition may be required.

By creating a binary decision matrix D, we are able to explicitly identify which subtasks need to be further decomposed as follows:

$$D[i][j] = \begin{cases} 1, & \text{if } S_{\text{similarity}}[i][j] < T_{\text{similarity}} \\ 0, & \text{otherwise} \end{cases} \quad (2)$$

Ultimately, by analyzing the decision matrix D, we are able to determine the degree of task decomposition, especially in scenarios with complex tasks, and this approach is very helpful in optimizing the decomposition of subtasks to ensure that tasks can be executed in the most efficient and accurate way. This cosine similarity-based approach not only has theoretical depth, but also shows strong results in practical applications, providing powerful tools and methods in the field of task decomposition and optimization.

Through this method, we provide feedback on incompletely decomposed subtasks and iteratively refine them until we generate the desired executable tasks for the robot. Ultimately, the task decomposition results are transmitted to the robot on the device side in the form of instructions. These instructions encapsulate robot executable functions and are transmitted to the edge's Large Language Model through a prompting mechanism, thereby transforming the task decomposition results into executable instructions for the robot on the device side.

## 4 EXPERIMENT

We designed an experiment to validate the feasibility of our proposed architecture, employing a quadruped robot and a drone for collaboration. In this experiment, we achieved the goal of policy generation and motion control for the quadruped robot in an environment without prior mapping. Leveraging user requirements and policies generated by the cloud-based large language model, we performed task decomposition using the visual model and linguistic model at the edge, enabling autonomous movement and task execution in an open environment.

As shown in Fig. 2, the user releases the task that the quadruped robot needs to perform and communicates the task requirements to the big language model at the edge. First, the large language model controls the UAV to collect environmental information through commands, and the UAV flies at a fixed point and transmits the collected environmental information to the edge computing layer, which is responsible for generating task instructions. The macrolanguage model combined with the vision model first performs task decomposition and then determines whether the task decomposition is complete through the cosine similarity, and then generates a sequence of task commands that can be executed by the quadruped robot through feedback iteration, which is then transmitted to the quadruped robot at the device end, so that the device can complete various actions in an orderly manner according to the requirements of the command sequence, and realize the complex motion control, so as to realize highly efficient human-computer interaction [11].

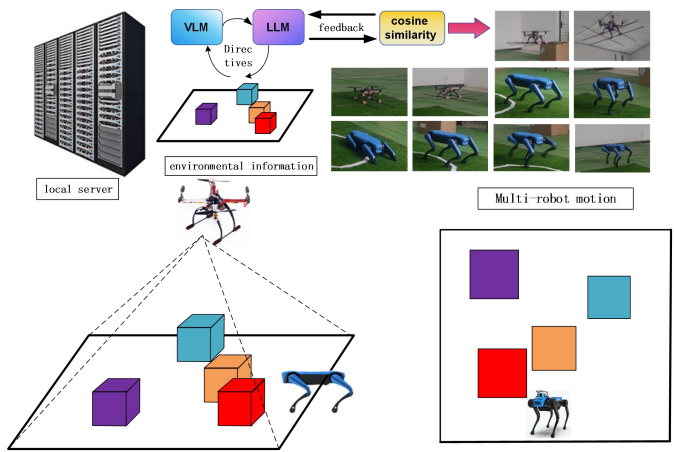

**Figure 2.** Experimental Schematic of Multi-Robot Strategy Generation and Motion Control

As shown in Fig. 3, we have accomplished the policy generation and motion control of the UAV and the quadruped robot in a realistic scenario, and the quadruped robot is able to complete the tasks proposed by the human according to the command sequences generated by the edge computing layer by providing the scenario environment information from the UAV and realizing the task execution in an open environment[12].

Our experiments achieved good results in reality, solved the problem of incomplete task decomposition of user requirements through large language modeling, and successfully realized the human-robot interaction between multiple robots and users in open scenarios, through which we verified the excellence of the architecture proposed in this paper in terms of multirobot task decomposition as well as multirobot strategy generation and motion control.

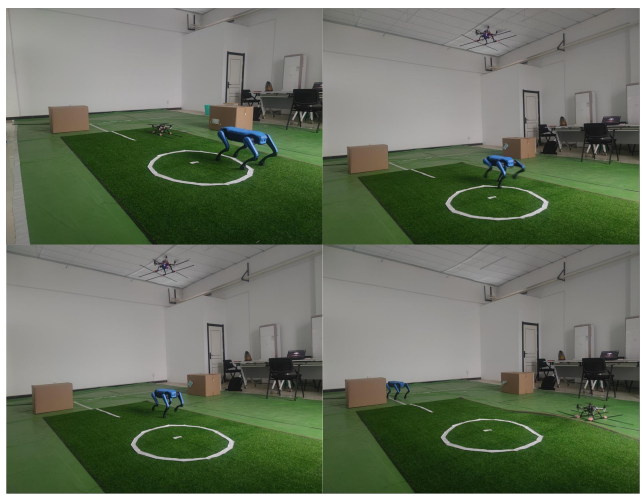

**Figure 3.** Multi-Robot Strategy Generation and Motion Control Experiments

## 5 CONCLUSION

In this paper, we propose an innovative cloud-edge-end architecture that fully utilizes the potential of large language models to address the problem of insufficient task decomposition in robotics. Our approach includes policy generation and task decomposition for tasks, as well as cosine similarity-based task evaluation. In the cloud, our large language model generates the robot's policy, while at the edge, the vision and language models work together to transform the policy into executable instructions for the robot. This architecture not only improves the efficiency of robots working together, but also ensures adequate task decomposition. Through our experiments, we demonstrate that this approach effectively improves the strategy generation and motion control of robots, bridging the gap between decision making and action execution.

**ACKNOWLEDGMENTS**

This work is supported by the National Natural Science Foundation of China (No. U21A20485).